\documentclass[10pt,twocolumn,letterpaper]{article}

\usepackage{cvpr}
\usepackage{times}
\usepackage{epsfig}
\usepackage{graphicx}
\usepackage{amsmath}
\usepackage{amssymb}
\usepackage{bm}

\usepackage[pagebackref=true,breaklinks=true,letterpaper=true,colorlinks,bookmarks=false]{hyperref}

\cvprfinalcopy 

\def\cvprPaperID{2083} 

\newcommand{\x}{\bm{x}}
\newcommand{\hx}{\bm{\widehat{x}}}
\newcommand{\I}{\mathcal{I}}

\newcommand{\A}{A}
\newcommand{\coh}{\mathcal{L}}
\newcommand{\D}{\bm{D}}
\newcommand{\sal}{S}
\newcommand{\map}{M}
\newcommand{\prior}{P}
\newcommand{\coef}{\bm{\alpha}}

\begin{document}

\title{A Reverse Hierarchy Model for Predicting Eye Fixations}

\author{Tianlin Shi\\
Institute of Interdisciplinary
Information Sciences\\
Tsinghua University,
Beijing 100084, China\\
{\tt\small tianlinshi@gmail.com}
\and
Ming Liang\\
School of Medicine\\
Tsinghua University,
Beijing
100084, China\\
{\tt\small liangm07@mails.tsinghua.edu.cn}
\and
Xiaolin Hu\\
Tsinghua National Laboratory for Information Science and
Technology (TNList)\\
Department of Computer Science and Technology, Tsinghua University,
Beijing
100084, China\\
{\tt\small xlhu@tsinghua.edu.cn}
}

\maketitle

\begin{abstract}
  A number of psychological and physiological evidences suggest that early visual attention works in a coarse-to-fine way, which lays a basis for the reverse hierarchy theory (RHT). This theory states that attention propagates from the top level of the visual hierarchy that processes gist and abstract information of input, to the bottom level that processes local details. Inspired by the theory, we develop a computational model for saliency detection in images. First, the original image is downsampled to different scales to constitute a pyramid. Then, saliency on each layer is obtained by image super-resolution reconstruction from the layer above, which is defined as unpredictability from this coarse-to-fine reconstruction. Finally, saliency on each layer of the pyramid is fused into stochastic fixations through a probabilistic model, where attention initiates from the top layer and propagates downward through the pyramid.  Extensive experiments on two standard eye-tracking datasets show that the proposed method can achieve competitive results with state-of-the-art models.
\end{abstract}

\section{Introduction}

\begin{figure}
\begin{center}
\includegraphics[width = 0.48 \textwidth]{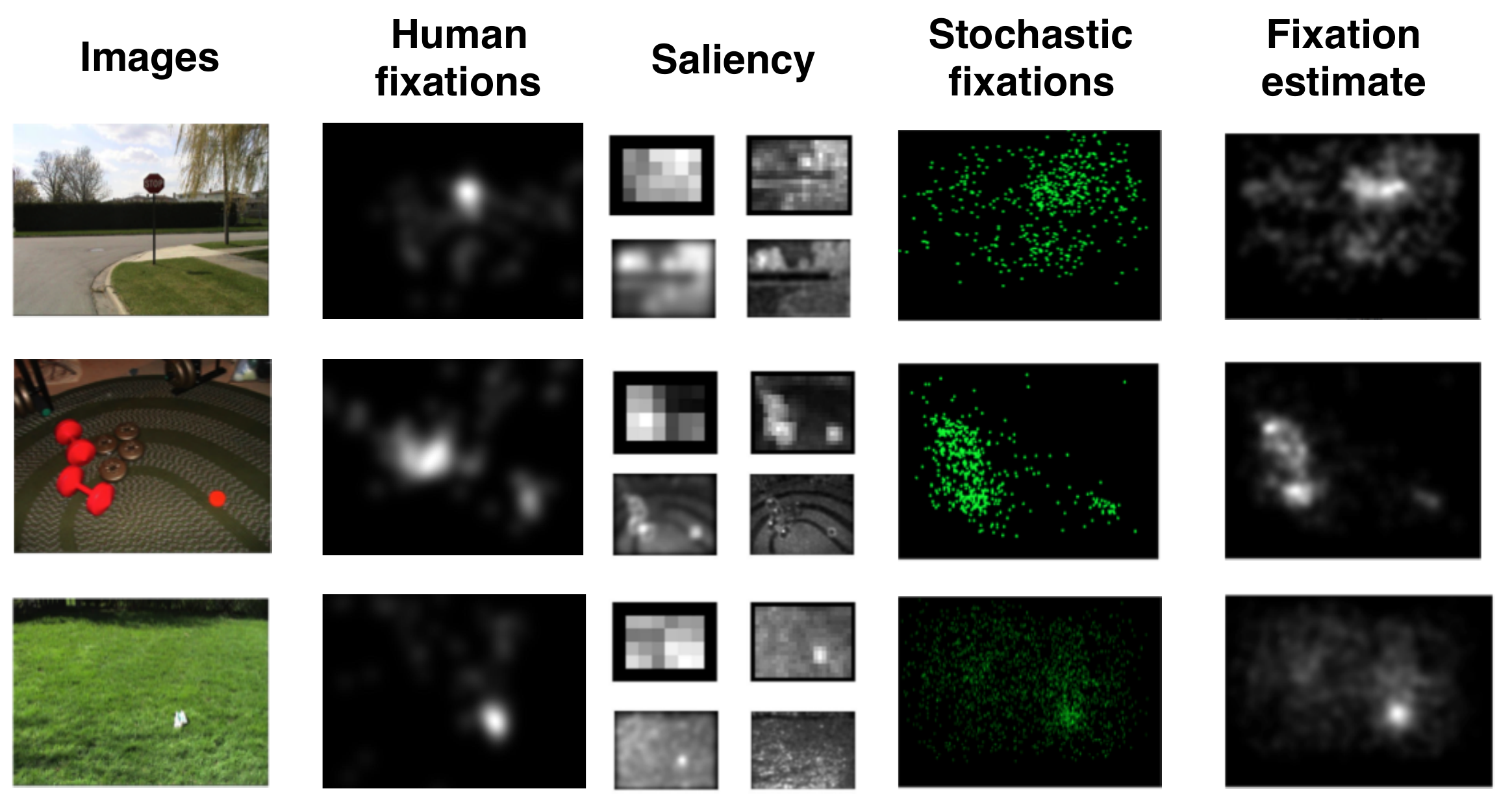}
\end{center}
\caption{Saliency computing and fusion in a reverse hierarchy scheme. \emph{Images}: taken from the TORONTO dataset~\cite{bruce2006saliency}. \emph{Human fixations:} eye fixations by human subjects in the eye-tracking experiment. \emph{Saliency:} saliency emerges from image super-resolution from four pairs of coarse-to-fine scales. \emph{Stochastic fixations: } fixations generated by the reverse hierarchy model. \emph{Fixation estimate: } blurred from the stochastic fixation map.}
\end{figure}

Human vision system can selectively direct eyes to informative and salient parts of natural scenes. This ability allows adaptive and efficient allocation of limited computational resources to important objects. Though enjoying great potential in various applications of computer vision, predicting eye fixations, however, remains  a challenging task. The underlying difficulty inherits from the ambiguous notion of what attracts eye fixations, or what is \emph{salient}. In fact, the theoretical investigation of visual saliency has aroused enduring controversies \cite{wolfe2003moving}. One possible explanation often adopted in the design of saliency detection approaches is the Feature Integration Theory (FIT) \cite{treisman1998perceiving}. According to FIT, attention serves as a mechanism to coherently combine features for the perception of objects. Therefore, starting from Itti and Koch ~\cite{itti1998model}, eye fixations are commonly predicted by directly conjoining saliency activations from multiple channels, which can be global and local channels~\cite{borji2012exploiting}, multiple features ~\cite{harel2007graph, garcia2012saliency} and so on.

Anatomical and physiological studies have shown that human visual system is organized hierarchically, which is believed to be advantageous in efficient processing of visual input. Computational studies have shown that hierarchical models (\eg HMAX~\cite{serre2007robust}, CDBN \cite{lee2009convolutional}) are effective for object recognition. Most saliency detection models, however, do not seriously take this into account. An obvious method to fill this gap is to develop hierarchical bottom-up models for saliency detection in the manner of HMAX, CDBN and the like. But there exists theoretical alternatives. The Reverse Hierarchy Theory (RHT)~\cite{hochstein} argues that parallel feedforward feature activation acts implicitly at first to construct a coarse gist of the scene, while explicit perception incrementally incorporates fine details via feedback control. This theory potentially has tremendous applications in computer vision including image segmentation, object recognition and scene understanding, however, computational studies are scarce. In this paper, we present an effective model based on RHT for saliency detection, which proves that RHT is helpful at least in this particular computer vision application. As for this application, a more direct evidence for the proposed model refers to a psychophysical study \cite{judd2011fixations} which showed that fixations from low-resolution images could predict fixations on higher-resolution images.

Our main idea is to model the coarse-to-fine dynamics of visual perception. We take a simple strategy to construct a visual hierarchy by inputting images at different layers with different scales, obtained by downsampling the original image. The higher layers receive coarser input and lower layers receive finer input. On each layer, saliency is defined as unpredictability in coarse-to-fine reconstruction through image super-resolution \cite{freeman2000learning, freeman2002example, yang2008image}.
The saliency on each layer is then fused into fixation estimate with a probabilistic model that mimics reverse propagation of attention. Throughout the paper, we call the proposed model a reverse hierarchy model (RHM).

The coarse-to-fine dynamics, however, is not the only property of RHT. In fact, RHT is closely related to the biased competition theory of attention~\cite{desimone1995neural, desimone1998visual}, which claims that attentional competition is biased by either stimulus-driven or task-dependent factors. Our model deals with fixation prediction in the free viewing task, which can be regarded as an implementation of the stimulus-driven bias. In addition, the image pyramid is a very coarse approximation of the highly complex structure of the visual hierarchy in the brain, which only utilizes the fact of increasing receptive field sizes along the hierarchy. Therefore, some closely related concepts to RHT, such as perceptual learning~\cite{ahissar2004reverse}, would not be discussed in the paper.


\section{Related Work}
\label{sec:related}

The majority of computational attention modeling studies follow the Feature Integration Theory~\cite{treisman1998perceiving}.  In particular, the pioneering work by Itti \etal \cite{itti1998model, itti2001computational} first explored the computational aspect of FIT by searching for center-surround patterns across multiple feature channels and image scales. This method was further extended  through integration of color contrast~\cite{le2006coherent}, symmetry~\cite{kootstra2008paying}, \etc. Random Center Surround Saliency \cite{vikram2012saliency} adopted a similar center-surround heuristic but with center size and region randomly sampled. Harel \etal~\cite{harel2007graph} introduced a graph-based model that treated feature maps as fully connected nodes, while the nodes communicated according to their dissimilarity and distance in a Markovian way. Saliency was activated as the equilibrium distribution.

Several saliency models adopted a probabilistic approach and modeled the statistics of image features. Itti and Baldi~\cite{itti2006bayesian} defined saliency as surprise that arised from the divergence of prior and posterior belief. SUN~\cite{zhang2008sun} was a Bayesian framework using natural statistics, in which bottom-up saliency was defined as self-information. Bruce and Tsotsos~\cite{bruce2006saliency} proposed an attention model based on information maximization of image patches. Garcia \etal~\cite{garcia2012saliency} defined the saliency by computing the Hotelling's T-squared statistics of each multi-scale feature channel. Gao \etal~\cite{gao2007discriminant} considered saliency in a discriminative setting by defining the KL-divergence between features and class labels.

A special class of saliency detection schemes was frequency-domain methods.  Hou and Zhang~\cite{hou2007saliency} proposed a spectral residual method, which defined saliency as irregularities in amplitude information. Guo \cite{guo2008spatio} explored the phase information in the frequency domain with a Quaternion Fourier Transform. Recently, Hou \etal \cite{hou2012image} introduced a simple image descriptor, based on which a competitive fast saliency detection algorithm was devised.

Different from our proposal, the conventional practice in fusing saliency at different image scales and feature channels was through linear combination. Borji \cite{borji2012exploiting} proposed a model that combined a global saliency model AIM~\cite{bruce2006saliency} and a local model~\cite{itti1998model, itti2001computational} through linear addition of normalized maps. Some models learned the linear combination weights for feature channels. Judd \etal~\cite{judd2009learning} trained a linear SVM from human eye fixation data to optimally combine the activation of several low-, mid- and high-level features. With a similar idea, Zhao and Koch~\cite{zhao2011learning} adopted a regression-based approach.


Our model is characterized by a top-down flow of information. But it differs from most existing saliency detection models that incorporate top-down components such as ~\cite{wolfe2004attributes,torralba2006contextual,zhang2008sun,liu2011learning} in two aspects. First, a biased prior (\eg, context clues, object features, task-related factors) is often needed in those models,  serving as the goal of top-down modulation, which is not necessary in our model. Second, hierarchical structure of the visual cortex is not considered in those models, but plays a significant role in our model.

Nevertheless, there were a few preliminary studies trying to make use of the hierarchical structure for saliency detection and attention modeling. The Selective Tuning Model~\cite{Tsotsos95} was such a model. It was a biologically plausible neural network that modeled visual attention as a forward winner-takes-all process among units in each visual layer. A recent study \cite{yan2013hierarchical} used hierarchical structure to combine multi-scale saliency, with a hierarchical inference procedure that enforces the saliency of a region to be consistent across different layers.

\section{Saliency from Image Super-Resolution}

\begin{figure}
\begin{center}
\includegraphics[width=0.47 \textwidth]{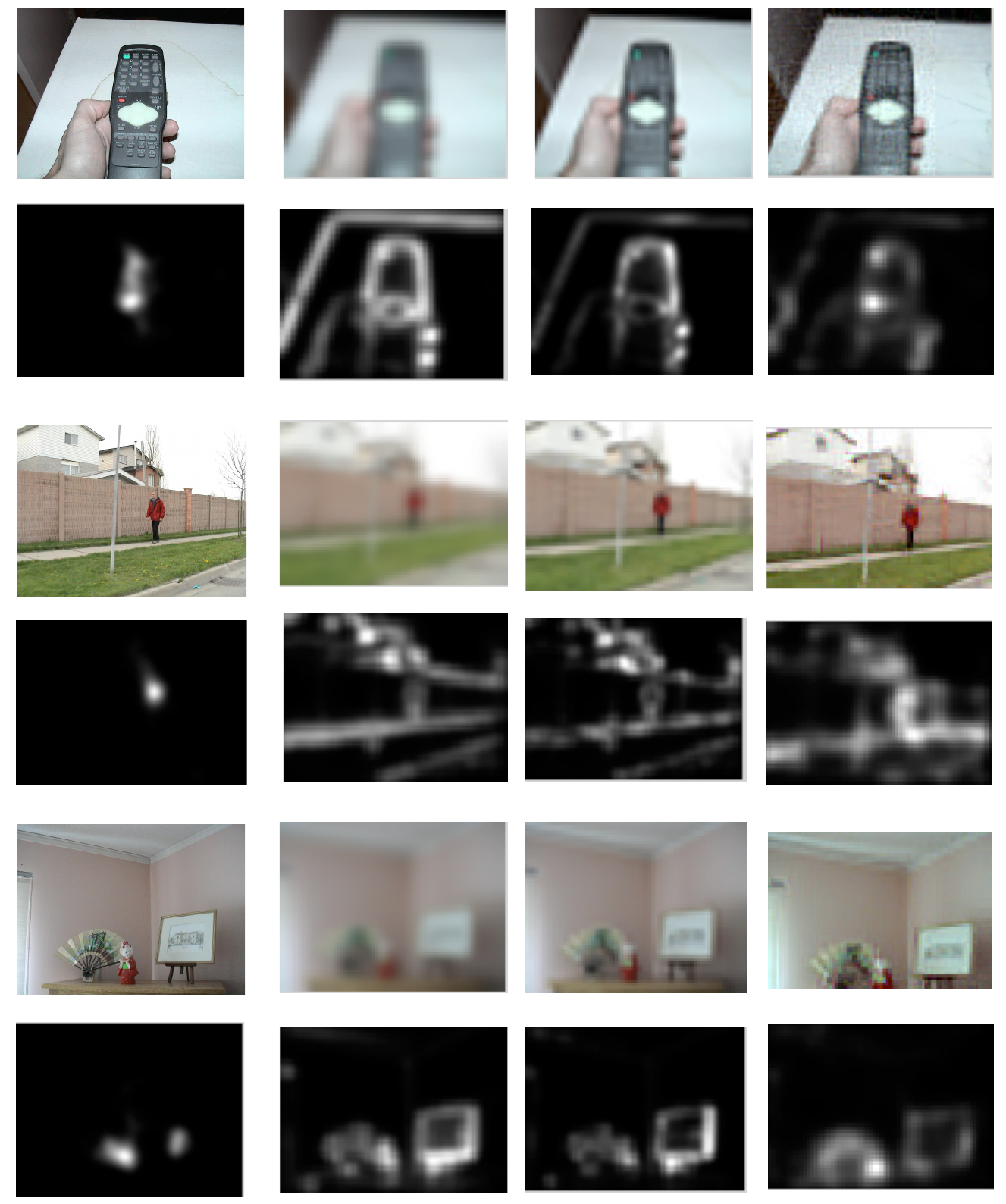}
\end{center}
\caption{Saliency from coarse-to-fine reconstruction.  The first column shows the original images and ground truth saliency. The second to the fourth columns show the reconstructed images by LS, BI and CS, respectively, together with predicted saliency. }
\label{fg:rec}
\end{figure}

In this section, a coarse-to-fine saliency model based on image super-resolution is presented. We consider an image at two consecutive scales in an image pyramid: a coarse one $\I_l$ and a fine one $\I_h$. Inspired by RHT, we define saliency as details in  $\I_h$ that are unpredictable from $\I_l$. In the next section, we discuss how to fuse saliency on each layer of the pyramid into fixation estimate.

\subsection{Saliency as Unpredictability}\label{subsec:unpredict}

Predicting $\I_h$ using the information of $\I_l$ is closely related to image super-resolution, which has been extensively studied using techniques including Markov random field~\cite{freeman2000learning}, example-based learning~\cite{freeman2002example}, compressive sensing~\cite{yang2008image}, \etc. In patch-based representation of images, the problem is to predict a high-resolution $H \times H$ patch $\x_h \in \I_h$ from its low-resolution $L \times L$ counterpart $\x_l \in \I_l$. For convenience of notation, we also use $\x_h$ and $\x_l$ as $H^2$ and $L^2$ dimensional vectors, which are computed by reshaping the corresponding patches. Then $\x_l$ is obtained by blurring and downsampling $\x_h$:
\begin{equation} \label{eq:sense}
\x_l = G B \x_h~,
\end{equation}
where $B$ denotes a $H^2 \times H^2$ blurring matrix (throughout the paper a Gaussian matrix is used) and $G$ represents a $L^2 \times H^2$ downsampling matrix. Let $\hx_h$ denote the reconstructed patch by some method $\mathcal{A}$, which summarizes the best knowledge one can recover from the coarse perception of $\x_h$ via $\x_l$. The reconstruction error of $\hx_h$ from $\x_h$ naturally represents the fine-scale information that cannot be recovered. Therefore, we define saliency $\sal(\x_h | \x_l)$  as the Normalized Mean Square Error (NMSE):
\begin{equation} \label{eq:sal}
\sal(\x_h | \x_l) = \frac{||\hx_h-\x_h||_2^2}{||\x_h||_2^2}.
\end{equation}
The mean squared error is normalized so that $S(\x_h | \x_l)$ is robust to variations of the patch energy $||\x_h||^2$.

\subsection{Coarse-to-Fine Reconstruction}\label{subsec:c2f}
\label{sec:ctf}

The reconstruction from the coarse scale subject to the constraint (\ref{eq:sense}) is actually not well-defined, since given a low-resolution patch $\x_l$, there exists an infinite number of possible high-resolution patches $\x_h$. To resolve this issue, the basic idea is to incorporate some prior knowledge, which inherits from the properties of natural images. In what follows we discuss several possible reconstruction schemes with increasingly sophisticated prior knowledge.
\vspace{5pt}

\textbf{Linear Reconstruction (LR).}~Consider a trivial case: the coarse patch $\x_l = B \x_h$ is just the blurred version and we do nothing but output $\hx_h = \x_l$. Therefore, no prior is used in this case. Saliency can be computed according to \eqref{eq:sal}. As shown in Fig. \ref{fg:rec}, this method assigns more saliency to patches containing many high-frequency components like edges and textures.

\textbf{Bicubic Interpolation (BI).}~If we reconstruct $\x_h$ using bicubic interpolation, then we utilize a smoothness prior in image interpolation. Although this approach concentrates less on edges than the linear reconstruction, its prediction is still far from the ground truth. See Fig. \ref{fg:rec}.

With LR or BI, the saliency computed in \eqref{eq:sal} is the normalized $l_2$-norm of the Laplacian pyramid. In addition, the two techniques can be used to implement the center-surround strategy adopted in some saliency models, \eg \cite{itti1998model}.

\textbf{Compressive Sensing (CS).}~We now consider a more sophisticated prior of image structure -- sparsity \cite{Hyvarinen09-nis}. According to this prior, any patch $\x_h$ of a high-resolution image can be sparsely approximated by a linear combination of items in a dictionary $\D_h$:
\begin{equation} \label{eq:sparse_dec}
\x_h \approx \D_h \coef,
\end{equation}
for some sparse coefficients $\coef$ that satisfies $||\coef||_0 \leq K$ for some small $K$. Assuming $\coef$ is sparse, the theory of compressive sensing states that $\coef$ can be recovered from sufficient measurements $\x_l = G B \x_h$ by solving the following optimization problem~\cite{donoho2006most}:
\begin{equation} \label{eq:sparse_dec}
\min~ {||\bm{\alpha}||_{0}}  ~~~\text{      subject to }~~~  {||\D_l \bm{\alpha}-\x_l||_2 \leq \epsilon},
\end{equation}
where $\D_l=G B \D_h$ denotes the blurred and downsampled dictionary $\D_h$ and $\epsilon$ is the allowed error tolerance.
This is hard to solve, and in practice the following relaxed problem is often solved \cite{donoho2006compressed, donoho2006most}:
\begin{equation} \label{eq:sparse_dec}
\min~ {||\bm{\alpha}||_{1}}  ~~~\text{      subject to }~~~  {||\D_l \bm{\alpha}-\x_l||_2 \leq \epsilon}.
\end{equation}
The coefficients $\coef$  are then used to reconstruct $\hx_h$ by
\begin{equation} \label{eq:rec}
\hx_h = \D_h \coef.
\end{equation}

Once we have obtained $\hx_h$, saliency of the image patch can be computed using (\ref{eq:sal}). Preliminary results in Fig. \ref{fg:rec} indicate that the saliency obtained by compressive sensing can largely differ from that obtained by LR and BI.

The dictionaries $\D_l$ and $\D_h$ are constructed as follows. For each scale of the image pyramid, we first uniformly sample raw patches  $\{\bm{d}_h^i\}_{i=1}^n$ of size $H \times H$ ($n \gg H^2$), and stack them into a high-resolution dictionary $\D_h = [\bm{d}_{h}^1, \bm{d}_{h}^2, ..., \bm{d}_{h}^n]$. Then we apply the blurring matrix $B$ and downsampling matrix $G$ to each $\bm{d}_h^i$ to obtain $\bm{d}_l^i = G B \bm{d}_{h}^i$. So $\D_l = [\bm{d}_{l}^1, \bm{d}_{l}^2, ..., \bm{d}_{l}^n]$ is the collection of corresponding low-resolution patches. The use of overcomplete raw patches for $\D_h$ and $\D_l$ has been shown effective for image super-resolution~\cite{yang2008image}.


\subsection{Saliency Map}
A saliency map $\map$ is obtained by collecting patch saliency defined in \eqref{eq:sal} over the entire image. First, calculate
\begin{equation}
\widetilde{M}[i,j] = \sal(\x_h[i,j] ~|~ \x_l[i,j]),
\end{equation}
where $\x_h[i,j]$ is the patch centered at pixel $(i,j)$ in the image and $\x_l[i,j]$ is its low-resolution version. Then $\widetilde{M}$  is blurred with a Gaussian filter \cite{hou2012image} and normalized to be between $[0,1]$ to yield the final saliency map $\map$. One should not confuse this Gaussian filter with $B$ in Sections \ref{subsec:unpredict} and \ref{subsec:c2f}.

\section{Reverse Propagation of Saliency}

Now, we present a method to transform the saliency maps at different scales into stochastic eye fixations on the original image. Based on RHT~\cite{hochstein}, a reverse propagation model is presented, where attention initiates from top level and propagates downward through the hierarchy.

\subsection{Generating Fixations} \label{sec:fix}

We model attention as random variables $\A_0, \A_1, ..., \A_n$ on saliency maps $\map_0, \map_1, ..., \map_n$, which are ordered in a coarse-to-fine scale hierarchy. Specifically, let $\Pr[\A_k = (i,j)]$ denote the probability for pixel $(i,j)$ attracting a fixation. To define this probability, we need to consider factors that influence the random variable $\A_k$. First of all, the saliency map $\map_k$ is an important factor. Pixels with higher values should receive more fixations. Second, according to RHT, attention starts from $\map_0$, and then gradually propagates down along the hierarchy. Therefore, $\A_k$ should  also depend on $A_{k-1}, ..., A_{0}$.
For simplicity, we assume that only $A_{k-1}$ has an influence on $A_{k}$ while $A_{k-2}, ..., A_{0}$ do not.

Based on these considerations, we define
\begin{equation} \label{eq:saliency_chain}
\Pr[A_{k} | M_{k},A_{k-1}, ..., A_{0}] = \Pr[A_{k} | M_{k},A_{k-1}],
\end{equation}
for $k=1,...,n$. A log-linear model is used for this conditional probability
\begin{equation} \label{eq:loglinear}
\begin{array}{l}
\Pr[A_{k} = (i,j)| M_{k},A_{k-1}] ~~~~~~~~~~~~~~ \\
~~~ \propto \exp\Big( \eta \map_k(i,j)+\lambda \coh(A_{k}, A_{k-1})\Big),
\end{array}
\end{equation}
where $\coh(A_k, A_{k-1})$ is a spatial coherence term, $\eta$ and $\lambda$ are two constants. The spatial coherence term restricts the fixated patches to be close in space. The motivation of introducing this term comes from the fact that the visual system  is more likely to amplify the response of neurons that is coherent with initial perception \cite{hochstein, newman1997neural}. To compute the term, we first convert the coordinate $\A_{k-1}$ into the corresponding coordinate $(u, v)$ in the saliency map just below it, \ie $\map_{k}$. Then compute
\begin{equation}
\coh(A_k, A_{k-1}) = -\Big((i-u)^2+(j-v)^2\Big).
\end{equation}
In other words, the farther away a patch $\x$ is from $A_{k-1}$, the less likely it would be attended by $A_k$. Therefore, for predicting the fixation probability of any patch in the current layer, the model makes a tradeoff  between the spatial coherence with previous attention and its current saliency value.

If we do not consider any prior on the top layer, $\Pr[\A_0]$ depends on the saliency map only 
\begin{equation} \label{eq:fixation}
\Pr[\A_0 = (i,j)] \propto \exp\Big(\eta \map_0[i,j]\Big).
\end{equation}

We can then generate fixations via an ancestral sampling procedure from the probability model. Specifically, we first sample fixation $A_0$ on map $\map_0$ according to \eqref{eq:fixation}, and then for $k=1,2,\ldots$ sample $A_k$ on map $\map_k$ given $A_{k-1}$ on the coarser scale according to \eqref{eq:loglinear}. Finally, we collect all samples on the finest scale, and use them as prediction of the eye fixations.

\begin{figure}
\includegraphics[scale=0.125]{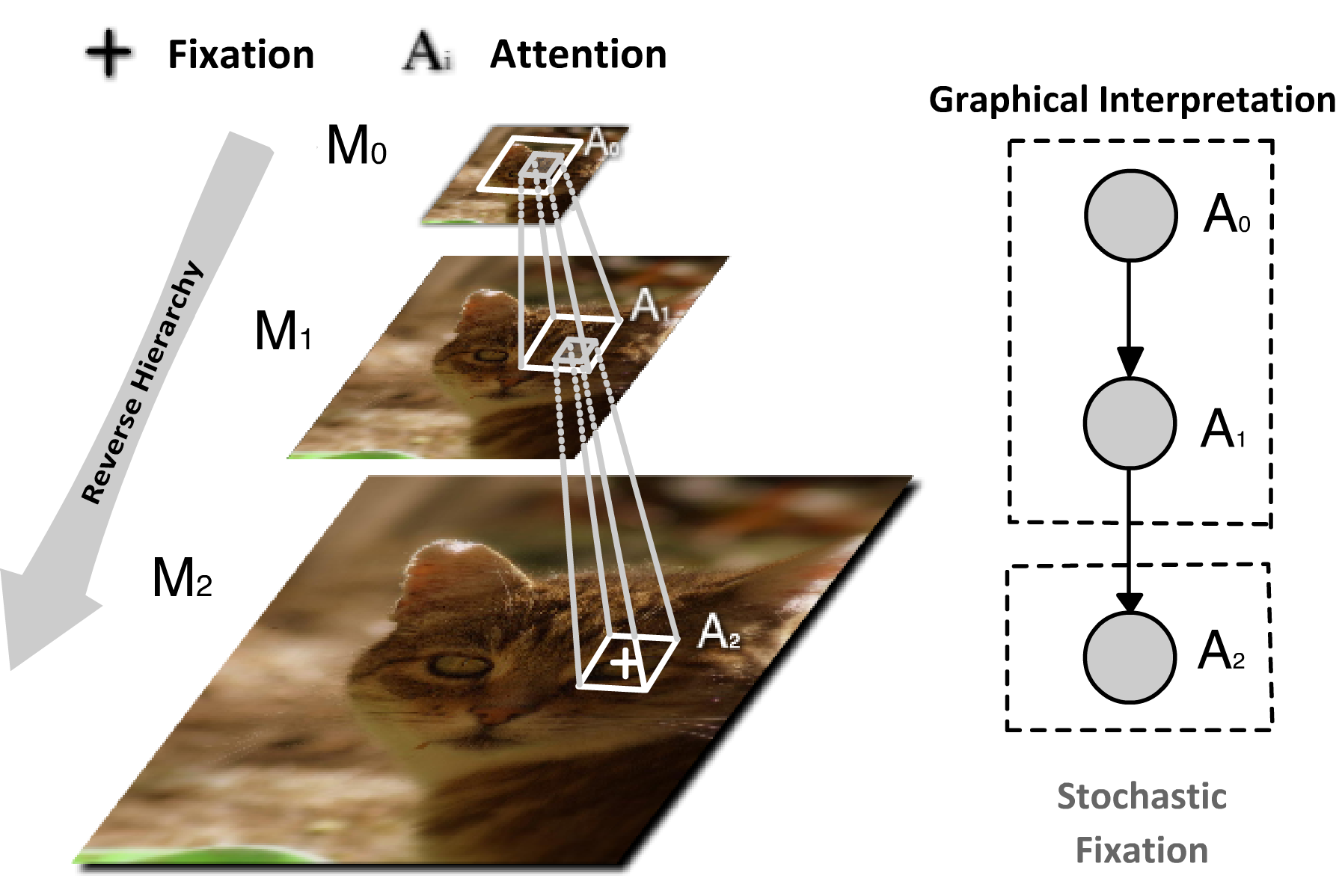}
\caption{Illustration of a three-layer reverse hierarchy model.  The attention initiates from a coarse image of a cat's face and propagates downward to lower-level details such as the cat's eyes.}
\end{figure}

\subsection{Incorporating Prior of Fixations}
\label{sec:prior}

 The proposed probabilistic model offers great flexibility for incorporating prior of fixations. 	This prior can be useful in capturing, for example, the top-down guidance of visual saliency from recognition~\cite{gao2009discriminant, yang2012top}, or central bias in eye-tracking experiments~\cite{tatler2005visual}. To achieve this, we extend the expression of $\Pr[A_0]$ as follows:
\begin{equation}\label{E:prior}
\Pr[\A_0 = (i,j)] \propto \exp\Big(\eta \map_0[i,j]+\theta \prior[i,j]\Big),
\end{equation}
where $P[i,j]$ encodes the prior information of pixel $(i,j)$ on the first map $M_0$ and $\theta$ is a weighting parameter.

For example, the central bias can be incorporated into the model by setting $P[i,j] =-[ (i-c_x)^2+(j-c_y)^2]$, where $(c_x, c_y)$ denotes the map center.

\section{Experiments}

\label{sec:exp}

\begin{figure*}
\begin{center}
\includegraphics[width=\textwidth]{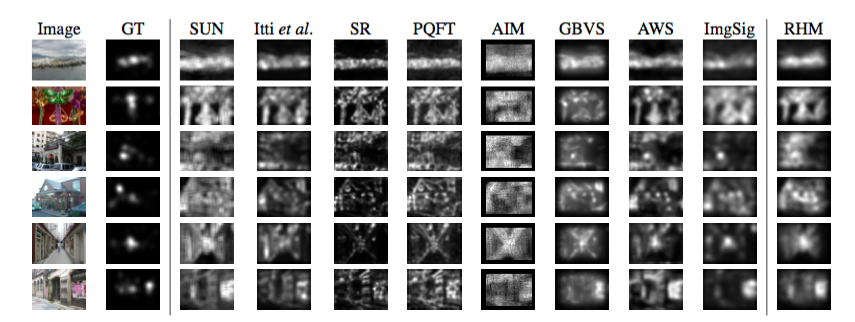}
\end{center}
\caption{Saliency maps produced by different models as well as the ground truth (GT).  The saliency maps of RHM were obtained from the predicted fixations blurred with a Gaussian filter.}
\label{figure:visual_compare}
\end{figure*}

\begin{table*}
\begin{center}
\begin{tabular}{r|ccc|ccc|ccc|ccc}
\hline
 & \multicolumn{6}{c|}{TORONTO Dataset \cite{bruce2006saliency}} & \multicolumn{6}{c}{MIT Dataset \cite{judd2009learning}} \\
\hline\hline
 & \multicolumn{3}{c|}{With Center} & \multicolumn{3}{c|}{Without Center}  & \multicolumn{3}{c|}{With Center} & \multicolumn{3}{c}{Without Center} \\
Model  & AUC   & NSS & S & AUC & NSS & S & AUC & NSS & S & AUC & NSS & S\\
Center & 0.801 & 1.118 & 0.472 & - & - &  - & 0.816 &  1.160 &  0.381 &  - & - & -\\
PQFT \cite{guo2008spatio}&    0.814& 1.466  &  0.483 &  0.751 & 0.362 & 0.372 &  0.801 & 1.033 & 0.361  & 0.722 &  0.645 & 0.275 \\
SR \cite{hou2007saliency}& 0.782  & 1.284 &  0.439 & 0.731 &  0.876 & 0.353 & 0.756 & 1.178 &  0.324 & 0.700 & 0.568 & 0.262 \\
SUN \cite{zhang2008sun} & 0.769  &   1.144 & 0.414 & 0.685 & 0.678 & 0.314 & 0.774 & 0.653 &  0.246 & 0.679 & 0.577 & 0.260 \\
AIM \cite{bruce2006saliency} & 0.815&  1.656 & 0.553 &  0.758& 0.960 & 0.404 & 0.750 & 0.782 &  0.414 &  0.739 & 0.569 & 0.291 \\
Itti \cite{itti2001computational} &  0.810  & 1.276  & 0.452  & 0.762 & 0.884 & 0.367  & 0.793 & 0.962 & 0.331 & 0.738 & 0.550 & 0.264 \\
GBVS \cite{harel2007graph} & 0.832    &  1.712  &  0.554 & 0.811 & 1.404 & 0.476 & 0.823 &   1.355 &  0.375 & 0.765 & 0.661 & 0.310\\
AWS \cite{garcia2009decorrelation} & 0.834    &  \textbf{1.803}  & 0.537  & 0.747  &   1.289 & 0.414 & 0.812 &  0.840 &  0.403 & 0.735 & 0.876 & 0.314 \\
ImgSig~\cite{hou2012image} & 0.840    &   1.746  & 0.548  & 0.802  & 1.509 & 0.478 & 0.823 &  0.898 &  0.376 &  0.761 & 0.790 & 0.311\\
\hline
\textbf{RHM} & \textbf{0.842} & 1.729 & \textbf{0.573} & \textbf{0.836} & \textbf{ 1.631}& \textbf{0.553} & \textbf{0.835} & \textbf{1.480} & \textbf{0.425} & \textbf{0.810} & \textbf{1.100} & \textbf{0.383}\\
\hline
\end{tabular}
\end{center}
\label{tb:auc}
\caption{Quantitative comparison of different models. }
\end{table*}

\subsection{Experiment Settings}

\textbf{Datasets.} The performance of the proposed reverse hierarchy model (RHM) was evaluated on two human eye-tracking datasets. One was the TORONTO dataset \cite{bruce2006saliency}. It contained 120 indoor and outdoor color images as well as fixation data from 20 subjects. The other was the MIT dataset \cite{judd2009learning}, which contained 1003 images collected from Flicker and LabelMe~\cite{russell2008labelme}. The fixation data was obtained from 15 subjects.

\textbf{Parameters.} The raw image $\mathcal{I}$ in RGB representation was downsampled by factors of 27, 9, 3 to construct a coarse-to-fine image pyramid. The patch size for super-resolution was set as $9 \times 9$ on each layer. To construct corresponding coarse patches, we used Gaussian blurring filter $B$ ($\sigma = 3$) and downsampling operator $G$ with a factor of 3. A total of 1000 image patches were randomly sampled from all images at the current scale to construct the dictionary $\D_h$, which is then blurred and downsampled to build $\D_l$.

In some experiments, we included a center bias \cite{tatler2005visual} in the model. This is achieved by switching $\theta$ from 0 to 1 in \eqref{E:prior}.

Note that the reverse propagation described in \eqref{eq:saliency_chain}-\eqref{eq:fixation} is a stochastic sampling procedure and we need to generate a large number of fixations to ensure unbiased sampling. We found that 20000 points on each image were enough to achieve good performance, which was adopted in all experiments. The stochastic points were then blurred with a Gaussian filter to yield the final saliency map. The standard deviation of the Gaussian filter was fixed as 4 pixels on saliency maps, which was about 5\% of the width of saliency maps of TORONTO images (TORONTO images had the same size), similar to \cite{hou2012image}.

\textbf{Evaluation metric.} Several metrics have been used to evaluate the performance of saliency models. We adopted Area Under Curve (AUC)~\cite{bruce2006saliency}, Normalized Scanpath Saliency (NSS)~\cite{peters2005components} and Similarity (S) \cite{judd2009learning}. Specifically, We used the AUC code from the GBVS toolbox \cite{harel2007graph}, NSS code from \cite{borji2012exploiting} and Similarity code from \cite{judd2009learning}. Following \cite{judd2009learning}, we first matched the histogram of the saliency map to that of the fixation map to equalize the amount of salient pixels in the map, and then used the matched saliency map for evaluation. Note that AUC was invariant to this histogram matching \cite{judd2009learning}.

\textbf{Models for comparison.} The proposed model was compared with several state-of-the-art models:
Itti \& Koch~\cite{itti2001computational}, Spectral Residual Methods (SR)~\cite{hou2007saliency}, Saliency based on Information Maximization (AIM)~\cite{bruce2006saliency}, Graph Based Visual Saliency (GBVS)~\cite{harel2007graph}, Image Signature (ImgSig)~\cite{hou2012image}, SUN framework~\cite{zhang2008sun} and Adaptive Whitening Saliency (AWS)~\cite{garcia2009decorrelation}. The implementation of these models were based on publicly available codes/software. Among these models, GBVS~\cite{harel2007graph}, ImgSig~\cite{hou2012image} and  AWS~\cite{garcia2009decorrelation} usually performed better than the others.

Inspired by the center bias~\cite{tatler2005visual}, we included a Center model as a baseline, which was simply a Gaussian function with mean at the center of the image and standard deviation being 1/4 of the image width~\cite{harel2007graph}. This simple model was also combined with other saliency detection models to account for the center bias, which could boost accuracy of fixation prediction. Following~\cite{harel2007graph}, this was achieved by multiplying the center model with the saliency maps obtained by these models in a point-wise manner.

\subsection{Results}

\begin{figure}
\begin{center}
\includegraphics[width = 0.45 \textwidth]{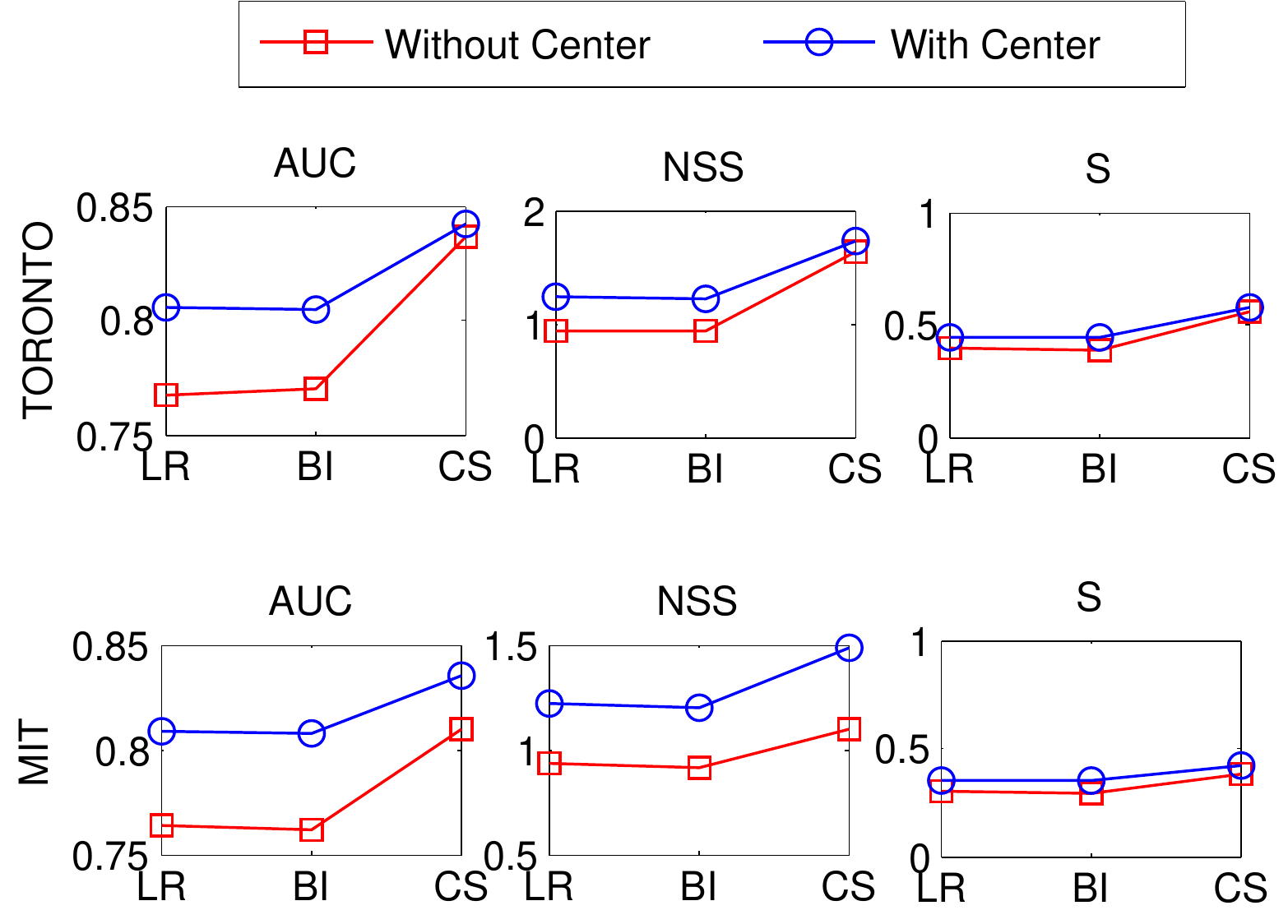}
\end{center}
\caption{Performance of RHM with three reconstruction methods LR, BI and CS. }
\label{fg:ctf}
\end{figure}

First, we compared different super-resolution techniques (LR, BI and CS) for eye fixation prediction. Fig. \ref{fg:ctf} shows the results of RHM with the three techniques. The CS method significantly outperformed LR and BI. Therefore, sparsity as a prior offers great advantage in discovering salient fine details. We then focused on RHM with CS in subsequent experiments.

Fig. \ref{figure:visual_compare} shows some qualitative comparison of the proposed model against existing models.
Table \ref{tb:auc} shows quantitative results under three metrics. As we can see, no single model could dominate others under all three metrics. However, in most cases (including both ``with" and ``without center" settings), the RHM outperformed the current state-of-the-art models. This demonstrated the reverse hierarchy theory as a promising way to predict human eye fixations.

\subsection{Contributions of Individual Components}

The RHM consists of two components: coarse-to-fine reconstruction (especially compressive sensing) and reverse propagation. Although the two components integrated together showed promising results, the contribution of each component to the performance is unclear. This is discussed as follows.

\textbf{Compressive sensing. } To identify the role of compressive sensing, we substituted it with other saliency models. Specifically, we replaced the saliency maps obtained from coarse-to-fine reconstruction by the saliency maps obtained by existing models. The models designed to work on a single scale, including SR~\cite{hou2007saliency}, AIM~\cite{bruce2006saliency},  SUN~\cite{zhang2008sun}, were applied to images of different scales to obtain multiple saliency maps. For multi-scale models such as Itti \& Koch~\cite{itti2001computational}, we use their intermediate single-scale results.

\begin{figure}
\begin{center}
\includegraphics[width = 0.5 \textwidth]{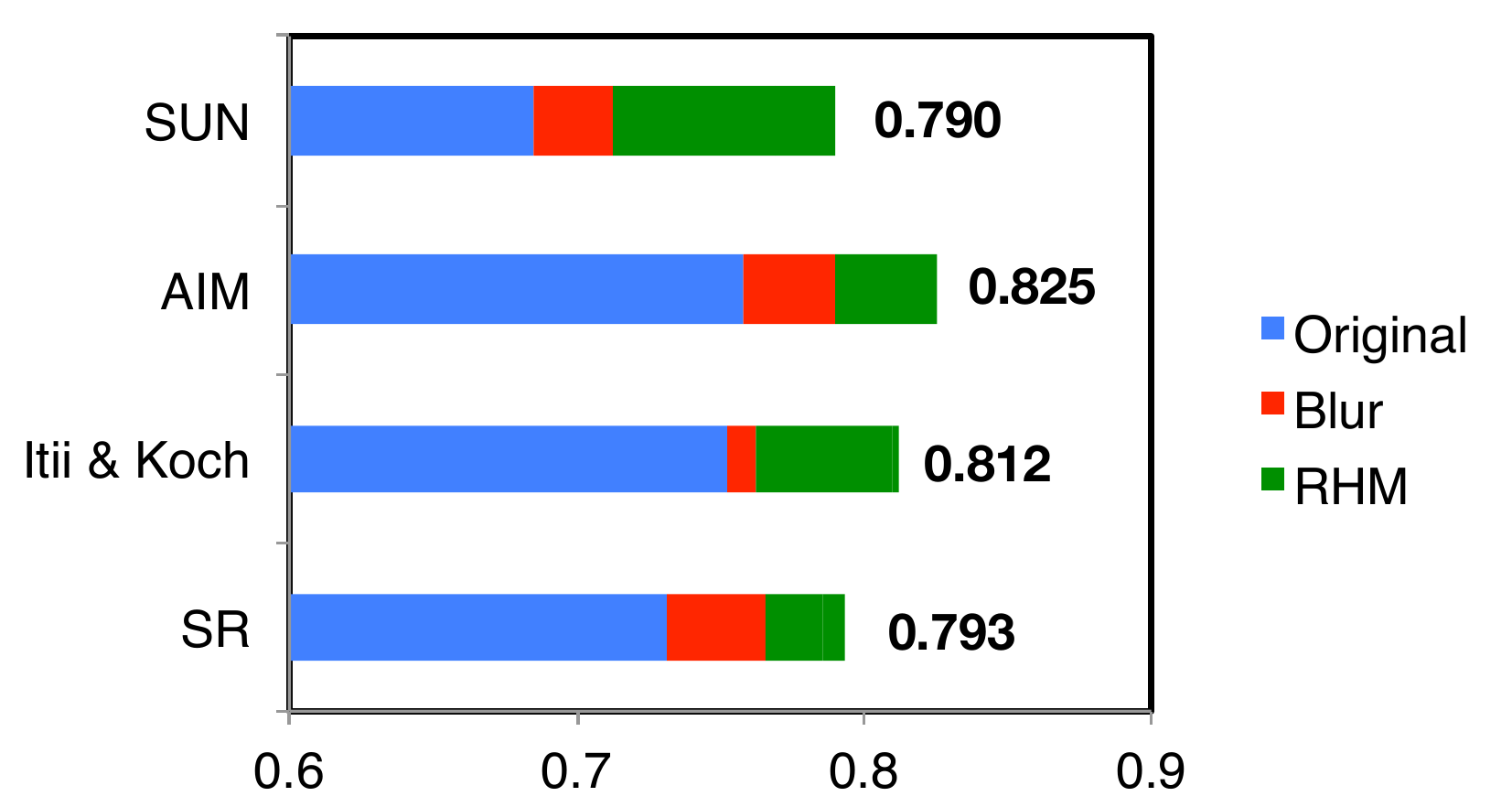}
\end{center}
\caption{Use RHM to boost the prediction accuracy of existing models on the TORONTO dataset.}
\label{fg:meta}
\end{figure}

Notice that blurring with a Gaussian filter is a necessary step in our model to obtain a smooth saliency map from stochastic fixations. Previous results have shown that blurring improved the performance of saliency models~\cite{hou2012image, borji2012exploiting}. For the sake of fairness, we also tested the models with the same amount of blurring (the sigma of Gaussian) used in RHM. Fig. \ref{fg:meta} shows the results on the TORONTO dataset. The reverse propagation procedure improved the AUC of these models. However, their performance is still behind RHM. Therefore, compressive sensing is a critical component in the RHM.

\textbf{Reverse propagation.} To investigate the effect of reverse propagation, we substituted it with linear combination of saliency maps, which is widely adopted in literature~\cite{itti2001computational, bruce2006saliency, borji2012exploiting}. Table \ref{tb:rp} shows the results. The linear combination produced an AUC between the best and worst that a single saliency map could achieve. However, RHM outperformed the best single-map performance. Therefore, through reverse propagation, RHM could integrate complementary information in each map for better prediction.

\begin{table}
\begin{center}
\begin{tabular}{c|ccccc}
\hline
Metric & $M_0$ & $M_1$ & $M_2$ & Linear & RHM \\
\hline\hline
AUC & 0.783 & 0.791 & 0.753 & 0.783 & 0.835 \\
NSS & 0.971 & 1.131 & 1.054 & 1.080 & 1.631 \\
S & 0.437 & 0.428 &  0.408 & 0.437 & 0.553 \\
\hline
\end{tabular}
\end{center}
\caption{AUC of single saliency maps and their linear combination (Linear) on the TORONTO dataset. The saliency maps $M_0$, $M_1$ and $M_2$ correspond to the downsampled images by factors of 27, 9 and 3 respectively. }
\label{tb:rp}
\end{table}

\section{Conclusion and Future Work}
\label{sec:conclusion}

In this paper, we present a novel reverse hierarchy model for predicting eye fixations based on a psychological theory, reverse hierarch theory (RHT). Saliency  is defined as unpredictability from coarse-to-fine image reconstruction, which is achieved by image super-resolution. Then a stochastic fixation model is presented, which propagates saliency from the top layer to the bottom layer to generate fixation estimate. Experiments on two benchmark eye-tracking datasets demonstrate the effectiveness of the model.

This work could be extended in several ways. First, it is worth exploring whether there exist better super-resolution techniques than compressive sensing for the proposed framework. Second, it is worth exploring if the ideas presented in the paper can be applied to a hierarchical structure consisting of different level of features, which play a significant role in the top-down modulation as suggested by RHT. Finally, in view of the similar hierarchical structure used in this study for saliency detection and other studies for object recognition, it would be interesting to devise a unified model for both tasks.

\section*{Acknowledgements}

This work was supported by the National Basic Research Program (973 Program) of China (Grant Nos. 2013CB329403 and 2012CB316301), National Natural Science Foundation of China (Grant No. 61273023, 91120011 and 61332007), Beijing Natural Science Foundation (Grant No. 4132046) and Tsinghua University Initiative Scientific Research Program (Grant No. 20121088071).


{\small
\bibliographystyle{ieee}
\bibliography{rhm}
}
\end{document}